\documentclass[sigconf]{acmart}

\usepackage{multirow}
\usepackage{caption}
\usepackage{subfigure}
\usepackage{balance}
\usepackage{bm}
\usepackage{hyperref}
\usepackage{booktabs}

\AtBeginDocument{%
  \providecommand\BibTeX{{%
    \normalfont B\kern-0.5em{\scshape i\kern-0.25em b}\kern-0.8em\TeX}}}

\copyrightyear{2020} 
\acmYear{2020} 
\setcopyright{acmcopyright}
\acmConference[MM '20]{Proceedings of the 28th ACM International Conference on Multimedia}{October 12--16, 2020}{Seattle, WA, USA}
\acmPrice{15.00}
\acmDOI{10.1145/3394171.3413668}
\acmISBN{978-1-4503-7988-5/20/10}

\begin{document}
\fancyhead{}

\title{Answer-Driven Visual State Estimator \\for Goal-Oriented Visual Dialogue}

\author{Zipeng Xu}
\email{xuzp@bupt.edu.cn}
\affiliation{\institution{Beijing University of Posts and Telecommunications}}
\author{Fangxiang Feng}
\email{fxfeng@bupt.edu.cn}
\affiliation{\institution{Beijing University of Posts and Telecommunications}}
\author{Xiaojie Wang}
\authornote{Corresponding author.}
\email{xjwang@bupt.edu.cn}
\affiliation{\institution{Beijing University of Posts and Telecommunications}}

\author{Yushu Yang}
\email{yangyushu@meituan.com}
\affiliation{\institution{Meituan-Dianping Group}}
\author{Huixing Jiang}
\email{jianghuixing@meituan.com}
\affiliation{\institution{Meituan-Dianping Group}}
\author{Zhongyuan Wang}
\email{wangzhongyuan02@meituan.com}
\affiliation{\institution{Meituan-Dianping Group}}

\begin{abstract}
A goal-oriented visual dialogue involves multi-turn interactions between two agents, Questioner and Oracle. During which, the answer given by Oracle is of great significance, as it provides golden response to what Questioner concerns. Based on the answer, Questioner updates its belief on target visual content and further raises another question. Notably, different answers drive into different visual beliefs and future questions. However, existing methods always indiscriminately encode answers after much longer questions, resulting in a weak utilization of answers. In this paper, we propose an Answer-Driven Visual State Estimator (ADVSE) to impose the effects of different answers on visual states. First, we propose an Answer-Driven Focusing Attention (ADFA) to capture the answer-driven effect on visual attention by sharpening question-related attention and adjusting it by answer-based logical operation at each turn. Then based on the focusing attention, we get the visual state estimation by Conditional Visual Information Fusion (CVIF), where overall information and difference information are fused conditioning on the question-answer state. We evaluate the proposed ADVSE to both question generator and guesser tasks on the large-scale GuessWhat?! dataset and achieve the state-of-the-art performances on both tasks. The qualitative results indicate that the ADVSE boosts the agent to generate highly efficient questions and obtains reliable visual attentions during the reasonable question generation and guess processes.
\end{abstract}

\ccsdesc[300]{Computing methodologies~Computer vision tasks}
\ccsdesc[300]{Computing methodologies~Discourse, dialogue and pragmatics}
\ccsdesc[300]{Computing methodologies~Natural language generation}
\ccsdesc[300]{Computing methodologies~Computer vision representations}
\keywords{Goal-Oriented Visual Dialogue; Attention Mechanism; Visual State Estimation}

\maketitle
\begin{figure}
  \centering
  \includegraphics[width=0.8\linewidth]{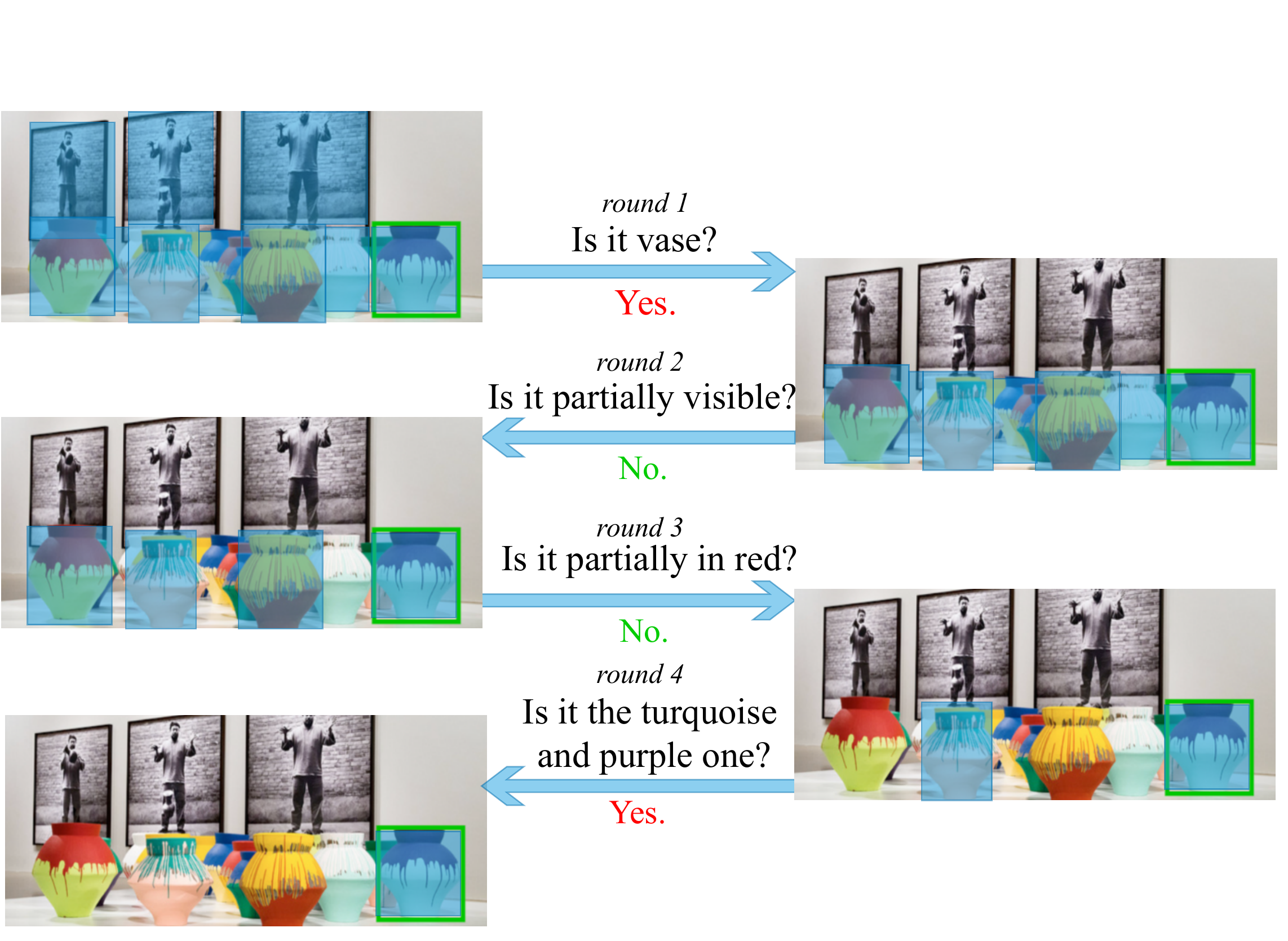}
  \caption{An example of discovering object in a image through dialogue. In which, answer largely determines the subsequent concerned visual information and question. Driven by successive questions and answers, the range of potential objects (highlighted in blue) shrinks and finally focuses on the target one (in green box).}
\end{figure}

\section{Introduction}

Goal-oriented Visual Dialogue, which means conducting multi-turn visual-grounded conversations with specific goals, is a comparatively new vision-language task while has attracted increased interests for its research significance and application prospect. As test-beds, image guessing tasks such as Guess-What \cite{de2017guesswhat} and Guess-Which \cite{das2017learning}, i.e. two-player games between Questioner and Oracle to retrieve visual content through dialogue, are proposed. In each round of the dialogue, the Questioner raises a visual-grounded question and gets respond from the Oracle (who predefines the visual target). After several rounds, Questioner is expected to make a right guess at the visual target. 

To conduct goal-oriented and vision-coherent dialogue, the AI agent should be able to learn a visual sensitive multimodal representation of the dialogue as well as a dialogue policy. Many works have been done on policy-learning. As Strub et al. \cite{Strub2017EndtoendOO} first introduce Reinforcement Learning (RL) to explore the dialogue strategy, later works take efforts on reward design \cite{Zhang_2018_ECCV, shukla-etal-2019-ask} or action selection \cite{DBLP:journals/corr/abs-1812-06398, Abbasnejad2018WhatsTK}. However, most of them employ a simple way to represent the multimodal dialogue by concatenating the two separately encoded modalities, i.e. language feature encoded by Recurrent Neural Network (RNN) and vision feature encoded by pre-trained Convolutional Neural Network (CNN). To improve the multimodal dialogue representation, various attention mechanisms have been proposed \cite{Zhuang2018ParallelAA, Deng2018VisualGV, Yang2019MakingHM}, where multimodal interactions are enhanced consequently. Although progresses have been made, unresolved issues still exist.

Firstly, none of the existing representation methods can distinguish among different answers in the dialogue history. The answer is always encoded right after the question without distinction. Since answer is usually a word of yes or no while question contains a longer word string, the effect of answer is relatively weak. However, in fact, answer largely determines the subsequent concerned visual information and question. As the object-discovery example in Figure 1, when the answer to the first question "Is it a vase?" is \emph{"yes"}, the questioner continues to pay attention to the vase and asks questions about the features that can best distinguish multiple vases; when the answer to the third question "Is it partially in red?" is \emph{"no"}, the questioner no longer pays attention to the vase in red and instead asks questions about the remaining candidates.

Secondly, the image in previous works is either encoded as a static embedding or attended by the dialogue history, which can hardly capture the influence of different answer on visual information. As mentioned above, different answer results in different concern changes on visual content. Generally, when answer is yes, we will focus on the question-related content for more detailed distinctive information within the confirmed candidates; when answer is no, we need to pay attention to the global area to find new possible candidates. Thus, a proper visual representation should have access to not only the global visual information but also the detailed distinctive information among candidates. Which kind of information is more important is dependent on the current question-answer (QA) state.

To address the above two issues, we propose an Answer-Driven Visual State Estimator (ADVSE), where the visual state is the QA-driven visual information dynamically updated through a dialogue. We formulate the ADVSE process in two steps. We firstly estimate the visual attention with Answer-Driven Focusing Attention (ADFA) and then accordingly estimate the visual state by Conditional Visual Information Fusion (CVIF). ADFA first uses a proposed sharpening operation to polarize the question-guided attention at current round, then inverts or maintains the attention based on different answers, and subsequently accumulates it in the final attention state. The effect of answer on the visual attention state is strengthened in this way. CVIF fuses overall information of the image and the difference information of the current focused candidate from other candidates under the guidance of the current QA, thus obtaining the estimated visual state. We apply ADVSE to build both Question Generator and Guesser for GuessWhat?!, where the specific goal is to discover an undisclosed object in a rich image scene. Experimental results show that both of them achieve state-of-the-art performances. 

To conclude, our main contributions are as follows. 
\begin{itemize}
\item First, we propose an Answer-Driven Visual State Estimator (ADVSE) to capture the influence of different answers in goal-oriented visual dialogue. 
\item Second, we apply the ADVSE to question generation and guess tasks on the large-scale GuessWhat?! dataset and achieve state-of-the-art performances on both tasks. 
\item Third, the qualitative results indicate that our ADVSE not only boosts the agent to generate highly efficient questions but also presents reliable visual attention during the reasonable question generation and guess processes.
\end{itemize}

\section{Related Works}

Goal-oriented dialogue requires the agent to complete a related task with a clear goal through multiturn conversations. Although goal-oriented spoken and text-based dialogues have been studied in Natural Language Processing committee for years \cite{10.1016/j.csl.2006.06.008, DBLP:journals/corr/BordesW16, DBLP:journals/corr/abs-1808-09996}, goal-oriented visual dialogue extends the setting to vision domain and is a relatively new and challenging field. Representatively, GuessWhat?! \cite{de2017guesswhat} aims to identify a predefined object in a real-world image through dialogue and GuessWhich \cite{das2017learning} is to figure out the referring image among various images. There are typically two dialogue agents, a Questioner and an Oracle, communicating together while the Questioner asks questions to figure out the undisclosed target and the Oracle, who predefines the target, responds accordingly. 

Question Generation is a core task in goal-oriented visual dialogue. De Vries et al. \cite{de2017guesswhat} first proposed a supervised model, where they extended the Hierarchical Recurrent Encoder Decoder (HRED) \cite{Serban2015BuildingED} by introducing the visual information, which is the image's FC8 feature obtained from a pre-trained VGG \cite{Simonyan15}. After that, various researches focused on dialogue policy learning. Strub et al. \cite{Strub2017EndtoendOO} introduced Reinforcement Learning (RL) to explore different dialogue strategy, which regarded question generation as a Markov Decision Process and used whether enabling a right guess as the reward function. Zhao et al. \cite{zhao2018learning} proposed a Temperature Policy Gradient method to make balance of exploration and exploitation while selecting words. Zhang et al. \cite{Zhang_2018_ECCV} designed a fine-grained reward mechanism based on the information provided by Oracle and Guesser. Some researchers explored the use of information uncertainty or changes to generate valuable questions \cite{lee2018answerer, Abbasnejad2018WhatsTK, shukla-etal-2019-ask}.

In these methods, the multimodal dialogue is encoded in the simplest way, where the CNN-encoded static image embedding is concatenated with the RNN-encoded changing dialogue history embedding to serve as the multimodal representation. However, encoding image as a static embedding is irrational, for the concerned image content changes as the dialogue progresses. Other than the simplest method, some attention-based methods are proposed to model the interaction between dialogue and image, computing dynamic visual information through dialogue. In PLAN network \cite{Zhuang2018ParallelAA}, the dialogue history embedding is jointly used with the image embedding to compute the attention on different regions, making it possible to provide dynamic visual information at each round. Deng et al. \cite{Deng2018VisualGV} proposed Accumulated Attention (A-ATT) mechanism that consists of three kinds of attention (query attention, image attention and objects attention), where the image is attended under the joint effect of dialogue history and object feature. Yang et al. \cite{Yang2019MakingHM} proposed a History-Aware Co-attention Network which includes two co-attention module, feature-wise co-attention module and element-wise co-attention module, while both of the attention are computed under the guidance of question and history feature. 

As we can see, none of the existing methods give special consideration to the effect of different answers. Most of the previous works weaken answers' effect by indiscriminately encoding the much shorter answers with a dialogue history encoder. On the contrary, the proposed Answer-Driven Visual State Estimator (ADVSE) explicitly exploits different answers in different ways to update the visual attention at each step and further fuses two types of visual information conditioning on different QA-state.

\section{Answer-Driven Visual State Estimator}

\begin{figure*}[ht]
  \centering
  \includegraphics[width=0.7\linewidth]{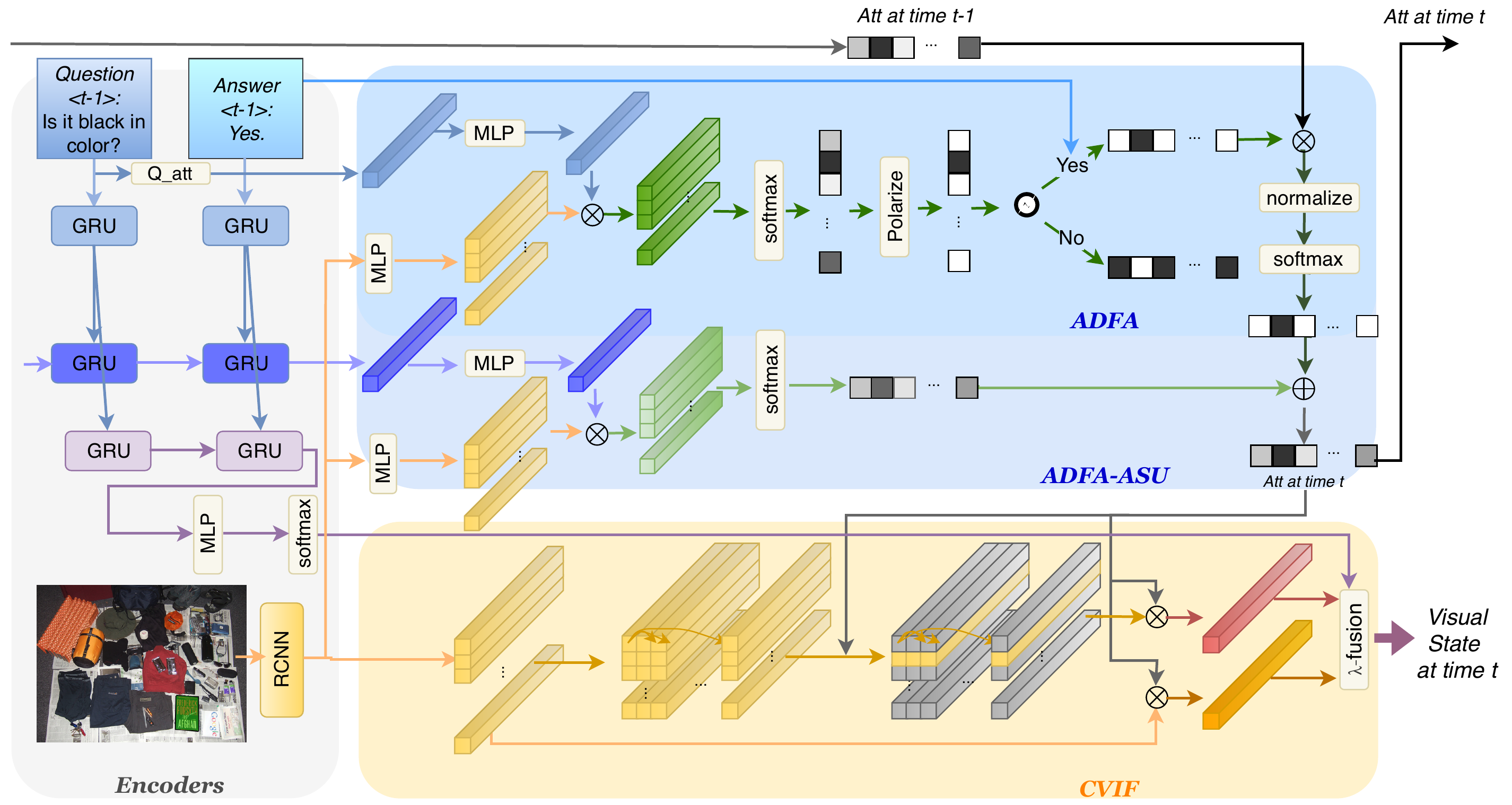}
  \caption{Block Diagram of the proposed Answer-Driven Visual State Estimator (ADVSE).}
\end{figure*}

This section introduces the proposed Answer-Driven Visual State Estimator (ADVSE). As in Figure 2, the estimator contains three parts, which are Encoders, ADFA-based Attention State Update (ADFA-ASU) and Conditional Visual Information Fusion (CVIF).

In the Encoders, visual information and language information are encoded separately. The ADFA-ASU estimates the visual attention while greatly considering the answer-driven effect with the proposed ADFA. Based on the estimated visual attention, the CVIF estimates the visual state by fusing the attended object overall information and the attended object difference information conditioning on the current QA state. They are introduced in detail below.

\subsection{Encoders}

{\bfseries Visual feature.} Given the input image \emph{I}, Faster-RCNN \cite{ren2015faster} is used to encode the image information. According to the static features provided by bottom-up attention \cite{anderson2018bottom}, the image representation is obtained:
\begin{equation}
  I = RCNN(image),
\end{equation}
of which, top-K region proposals are selected from each image. Here, K is simply fixed as 36, i.e. 
\begin{math}
  I = \{i_1, i_2, \ldots, i_{36}\} \in R^{36 \times 2048}
\end{math}.

{\bfseries Language feature.} Given the t rounds dialogue history $H=\{(q_1,a_1 ),\ldots,(q_t,a_t)\}$, where $q_t$ is the t-th round question and $a_t$ is the t-th round answer, a 2-layer GRU is applied to encode the dialogue. In concrete, the t-th round question $q_t$, which includes m words and whose word embeddings are $\{w_{t,1}^q$, \ldots, $w_{t,m}^q\}$, is encoded by $GRU^w$:
\begin{equation}
  h_{t,i}^q=GRU^w (w_{t,i}^q,h_{t,i-1}^q ).
\end{equation}

We use the last hidden state $Q_t=h_{t,m}^q$ as the representation of the question.

Similarly, the representation of current answer $A_t$ can be obtained. By feeding $Q_t$ and $A_t$ to the upper layer $GRU^c$, the representation of t-th round dialogue history $H_t$ is obtained:
\begin{equation}
  H_t=GRU^c([Q_t; A_t],H_{t-1}).
\end{equation}

\subsection{ADFA-ASU}

During the visual dialogue process, the attention state to the image dynamically updates, driven by the dialogue history and the current QA. In this section, we formulate the attention updating process by the proposed ADFA-ASU. At \emph{t}-th round, the attention state $att_t$ is updated by two parts: current QA caused Answer-Driven Focusing Attention (ADFA) ${att}_t^q$ and history guided attention ${att}_t^h$. The concrete modeling of ${att}_t^q$ and ${att}_t^h$ are described below:

Firstly, in Answer-Driven Focusing Attention (ADFA), the current turn QA-guided focusing attention state ${att}_t^q$ is modeled by the following four steps:

Step 1, calculate the question-guided image attention ${\alpha}_t^q$ according to Eq. 4-7:
\begin{equation}
  Q_t^m = \{h_{t,i}^q\}_{i=1}^m,
\end{equation}
\begin{equation}
  \widetilde{Q}_t^k = softmax(Q_t^m W_q^k)^T \odot Q_t^m,~\widetilde{Q}_t = [\widetilde{Q}_t^1; \widetilde{Q}_t^2],
\end{equation}
\begin{equation}
  F_t^q= W_Q \widetilde{Q}_t \odot W_I^q I,
\end{equation}
\begin{equation}
  \alpha_t^q= Softmax(W_F^q F_t^q + g).
\end{equation}

In order to extract the important information within a question, a 2-glimpse attention is utilized to extract the current question feature $\widetilde{Q}_t$ as in Eq. 4-5. The textual question feature and visual feature is then fused by Hadamard Product (Eq. 6). To enable end-to-end training with the subsequent discrete decision, we introduce Gumbel-Softmax sampler \cite{gumbel1948statistical} as well as the Gumbel-Softmax training trick \cite{jang2016categorical, Maddison2017TheCD} to compute the attention distribution as in Eq. 7. In concrete, we add $g$ (i.i.d. samples from Gumbel distribution) before the softmax activation during the training stage.

Step 2, polarize the $\alpha_t^q$ by a sharpening operation as shown in Eq. 8-9 to figure out the question-correlated objects:
\begin{equation}
  norm(\alpha_t^q) = \frac{\alpha_t^q - min(\alpha_t^q)}{max(\alpha_t^q)-min(\alpha_t^q)},
\end{equation}
\begin{equation}
  P(\alpha_{t,k}^q) = \left\{
  \begin{aligned}
  1 & , & if norm(\alpha_{t,k}^q) > \gamma, \\
  0 & , & else.
  \end{aligned}
  \right.
\end{equation}

The attention sharpening operation project the attention weight of each block $\alpha_{t,k}^q \in \alpha_t^q \in (0,1)$ into a binary value $P(\alpha_{t,k}^q) \in \{0,1\}$. It first applies the max-min normalization to $\alpha_t^q$ and gets $norm(\alpha_t^q)$ (Eq. 8), then filters the normalized attention by a threshold $\gamma$ (i.e. a hyperparameter) to get the polarized value $P(\alpha_{t,k}^q)$ (Eq. 9), which represents whether the object $i_k$ correlates to what $q_t$ asks.

Step 3, based on $P(\alpha_{t}^q)$, the answer to $q_t$ is used to determine the direction of the attention mask $M_t^q$ as shown in Eq. 10:
\begin{equation}
  M_t^q = \left\{
  \begin{aligned}
  P(\alpha_t^q) & , & if a_t == YES, \\
  \bm{1} - P(\alpha_t^q) & , & if a_t == NO, \\
  \bm{1} &, & otherwise.
  \end{aligned}
  \right.
\end{equation}

If the answer is \emph{“yes”}, the attention mask is $P(\alpha_t^q)$, which means the agent will hold attention on the currently concerned objects. The agent will keep paying close attention to the objects with the $P(.)$ of 1 and keep paying no attention to those objects with the $P(.)$ of 0. If the answer is \emph{“no”}, the attention mask is $1 - P(\alpha_t^q)$, which means the attentions on objects is going to be reversed. The agent will transfer its attentions to other objects whose $P(.)$ is 0 and no longer concern the objects whose $P(.)$ is 1 as they are denied by the Oracle. Otherwise, if the answer is \emph{"N/A"}, the $att_t^q$ will be kept unchanged, which is achieved by letting the elements in $M_t^q$ be 1 for all candidates. In this way, the answer plays a key role on forming the subsequent visual attention and therefore affects the visual state. 

Step 4, after calculating the influence of current round of question and answer, we update the focusing attention state $att_t^q$. The obtained attention mask $M_t^q$ is applied on the previous attention state $att_{t-1}$ by a Hadamard Product, and is then normalized. A learnable parameter $\tau$ and masked softmax are utilized to adjust the updated $att_t^q$ as shown in Eq. 11:

\begin{equation}
  {att}_t^q = maskedSoftmax(\frac{Norm(M_t^q \odot att_{t-1} )}{\tau}).
\end{equation}

Secondly, the history guided attention is calculated as follows:
\begin{equation}
  att_t^h = softmax(W_F^H (W_t^H H_t \odot W_I^H I)).
\end{equation}

Finally, we get the estimated attention state ${att}_t$ by adding ${att}_t^q$ and ${att}_t^h$:
\begin{equation}
  {att}_t = {att}_t^h + {att}_t^q
\end{equation}

The attention state is dynamically updated and gradually focused in this way as successive QA pair generates.

\subsection{CVIF}

In CVIF, we firstly compute the attended difference information $D_{att}^t$ and the attended overall information $I_{att}^t$ based on the attention state estimated in ADFA-ASU. Finally, we fuse the two types of visual information conditioning on the current QA to obtain the current visual state estimation $V_t$.

First, the difference information between the mostly focused object and others is achieved in two steps as follows:

Step 1, select the mostly focused object $i_{{selected}^t}$ according to the ${att}_t$:
\begin{equation}
  {selected}^t=argmax({att}_t).
\end{equation}

Step 2, compute the difference between $i_{{selected}^t}$ and other object, and then get the focused difference information guided by ${att}_t$, as described by the following formulas:
\begin{equation}
  D_{selected}^t=\{i_{{selected}^t}-i_j \}_{j=1}^N, 
\end{equation}
\begin{equation}
  D_{att}^t=D_{selected}^t \otimes {att}_t. 
\end{equation}

Then, the overall feature is calculated by:
\begin{equation}
  I_{att}^t=I \otimes {att}_t.
\end{equation}

Finally, $D_{att}^t$ and $I_{att}^t$ are fused conditioning on current QA-pair. The QA pair is first encoded as shown in Eq. 18, and then normalized by softmax to obtain the conditioning factor $\lambda_t$ as shown in Eq. 19. Then the estimated visual state $V_t$ is obtained by weighted summing the $D_{att}^t$ and $I_{att}^t$ with the factor $\lambda_t$, as shown in Eq. 20. 
\begin{equation}
  h_{t,q}^p=GRU^p (q_t,h_0),~P_t= GRU^p (h_{t,q}^p,a_t),
\end{equation}
\begin{equation}
  (\lambda_t,1-\lambda_t )=softmax(W_p P_t),
\end{equation}
\begin{equation}
  V_t=\lambda_t \odot D_{att}^t+(1-\lambda_t ) \odot I_{att}^t.
\end{equation}

Visual state estimation is a soft fusion of difference information and overall information conditioned on current QA-pair, which strengthens again the influence of current answer. 

\section{Using ADVSE for QGen and Guesser}

ADVSE is a general framework for goal-oriented visual dialogue. In this section, we apply it to model the Question Generator (QGen) and Guesser in GuessWhat?! game. We firstly combine the ADVSE with an ordinary hierarchical history encoder to get the multimodal dialogue representation:
\begin{equation}
  F_t = tanh(W_f[H_t;V_t]).
\end{equation}
In which, $H_t$ is the encoding of dialogue history (as in Eq. 3) and $V_t$ is the visual state estimated by ADVSE (as in Eq. 20). They are concatenated and then projected by an MLP to get the multimodal dialogue representation $F_t$.

On the basis of $F_t$, the ADVSE-QGen and ADVSE-Guesser are introduced as follows.

\begin{figure}[t]
\subfigure[ADVSE-QGen]{
\begin{minipage}[t]{0.8\linewidth}
\includegraphics[width=1\linewidth]{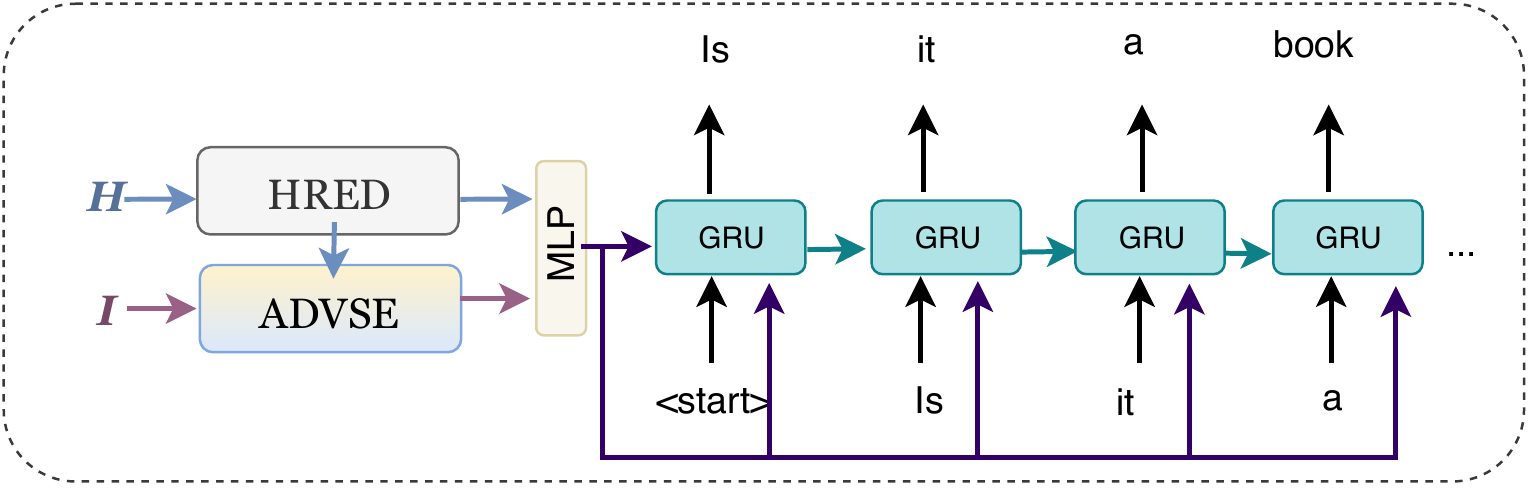}
\end{minipage}
}

\subfigure[ADVSE-Guesser]{
\begin{minipage}[t]{0.8\linewidth}
\includegraphics[width=1\linewidth]{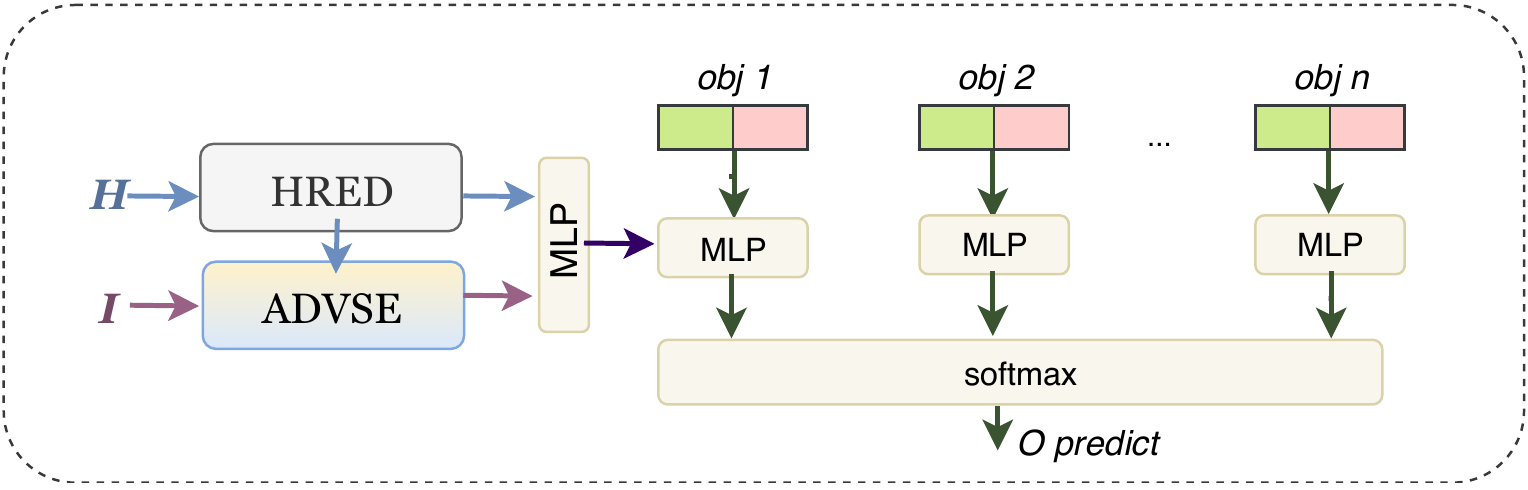}
\end{minipage}
}
\caption{Using ADVSE for QGen and Guesser.}
\end{figure}

\subsection{ADVSE-QGen Model}
In the process of visual dialogue, given image \emph{I}, dialogue history $H=\{(q_1,a_1 ),\ldots,(q_t,a_t)\}$, the QGen model needs to generate new question $q_{t+1}=(w_0^{t+1},w_1^{t+1},…,w_m^{t+1} )$, so as to get more information of the target object. As shown in Figure 3(a), the ADVSE-QGen Model is mainly modeled by ADVSE, HRED and a decoder.

Specifically, after ADVSE-based multimodal representation $F_t$ is obtained, the decoder takes $F_t$ as the initial incentive:
\begin{equation}
  h_{dec}^{t,0}=F_t.  
\end{equation}

The decoder is a single-layer GRU. The word vector is concatenated with the visual state estimation $V_t$ and the previous state, then used as the input at current state to predict the next word:
\begin{equation}
  w_i^{t+1}=GRU_d ([w_{i-1}^{t+1};V_t ],h_{dec}^{t,i-1} ).
\end{equation}
When stop token appears, the generation ends.

\subsection{ADVSE-Guesser Model}
With image \emph{I} and completed dialogue history $H=\{(q_1,a_1 ),\ldots,(q_T,a_T)\}$ in hands, a Guesser is expected to select the target object $o^*$ from the candidates $O=\{o_1,o_2,\ldots,o_n \}$ while it has access to the spatial information $s_O$ and the category information $c_O$ in addition. The ADVSE-Guess Model is mainly modeled by ADVSE, HRED and a classifier as shown in Figure 3(b).

The classifier first encodes the object representation $r_O$ from its category and spatial information as in Eq. 24, which is the same as the previous models \cite{Strub2017EndtoendOO}. 
\begin{equation}
  r_O = ReLU(W_{o}^2ReLU(W_o^1[s_O; c_O])).
\end{equation}

Then, softmax function is applied on the dot product between $F_T$ and $r_O$ to get the probability distribution. At last, the one with the maximum probability is selected:
\begin{equation}
  o_{predict}=argmax(softmax({F_T}^{'} r_O )).
\end{equation}

\section{Experiments}

We evaluate the models on the GuessWhat?! dataset, which has 155,281 dialogues based on 66,537 images, containing 134,074 different objects. There are 821,955 question-answer pairs in the dataset while the vocabulary size is 4900. We use standard dataset split (train set, validate set, test set).

In this section, we firstly report experimental results of ADVSE-QGen and ADVSE-Guesser respectively. We introduce the training details and evaluation metric, make comparisons with the state-of-the-art models and provide qualitative results. To verify the contribution of each component under different tasks, we conduct ablation study on both ADVSE-QGen and ADVSE-Guesser. Further, we report the experimental results of jointly using ADVSE-QGen and ADVSE-Guesser. The codes of our models are available at \href{https://github.com/zipengxuc/ADVSE-GuessWhat}{https://github.com/zipengxuc/ADVSE-GuessWhat}. 

\subsection{ADVSE-QGen}

\subsubsection{Training Details}

The QGen model is firstly trained in supervised way, and then trained by reinforcement learning.

In supervised learning, we minimized the negative likelihood loss. We use Adam \cite{Kingma2014AdamAM} with an initial learning rate of 1e-3, a batch size of 64 to train the QGen model for 20 epochs. Learning rate is decayed by 0.9 per epoch. The hyperparameter $\gamma$ in Sharpening Operation is set as 0.7.

Further, we train the model using the same reinforcement learning method as the baseline model \cite{Strub2017EndtoendOO}, where the QGen is modeled as a Markov Decision Process and uses the 0-1 reward that depends on whether a right guess can be made. We use Stochastic Gradient Descent (SGD) to train the model for 500 epochs with a learning rate of 1e-3 and a batch size of 64. We set the maximum round T = 8, the maximum length of each sentence m = 12. We use the same standard Oracle and Guesser as \cite{Strub2017EndtoendOO} while the trained benchmark Oracle and Guesser's errors on the test set are 21.9\% and 35.9\%, respectively.

\begin{table*}[ht]
  \caption{A comparison results of QGen on the task success rate evaluated by two types of Guesser, i.e. the standard Guesser\cite{Strub2017EndtoendOO} and the proposed ADVSE-Guesser. The upper part shows the results in SL while the bottom part shows the results in RL.}
  \label{tab:commands}
  \begin{tabular}{c|l|cccccccc}
    \toprule
    \multirow{2}{*}{ } & \multirow{2}{*}{\bf Approach} & \multicolumn{4}{c}{\bf (\%)New object}      & \multicolumn{4}{c}{\bf (\%)New game}    \\
    & & \bf Sampling      & \bf Greedy      & \bf Beam-search      & \bf Best   & \bf Sampling      & \bf Greedy      & \bf Beam-Search      & \bf Best   \\ \hline
    \multirow{4}{*}{Guesser\cite{Strub2017EndtoendOO}} & SL & 41.6 & 43.5 & 47.1 & 47.1 & 39.2 & 40.8 & 44.6 & 44.6 \\ 
    & DM                        & - & - & - & - & - & - & - & 42.19 \\
    & VDST-SL                        & 45.02 & 49.49 & - & 49.49 & 44.24 & 45.94 & - & 45.94 \\ 
    & \bf ADVSE-QGen                        & \bf 47.55 & \bf 50.66 & \bf 47.47 & \bf 50.66 & \bf 44.75 & \bf 47.03 & \bf 44.70 & \bf 47.03 \\ 
    \hline
     ADVSE-Guesser & \bf ADVSE-QGen                     & \bf 48.01 & \bf 54.06 & \bf 50.66 & \bf 54.06 & \bf 46.32 & \bf 50.94 & \bf 47.89 & \bf 50.94 \\ 
    \midrule[0.75pt]
    \multirow{6}{*}{Guesser\cite{Strub2017EndtoendOO}} & RL & 62.8 & 58.2 & 53.9 & 62.8 & 60.8 & 56.3 & 52.0 & 60.8 \\ 
    & VQG                        & 63.2 & 63.6 & 63.9 & 63.9 & 59.8 & 60.7 & 60.8 & 60.8 \\ 
    & Bayesian      & 61.4 & 62.1 & 63.6 & 63.6 & 59.0 & 59.8 & 60.6 & 60.6 \\ 
    & GDSE-C                     & - & - & - & 63.3 & - & - & - & 60.7 \\ 
    & ISM                        & - & 64.2 & - & 64.2 & - & 62.1 & - & 62.1 \\ 
    & TPG                        & - & - & - & - & - & - & - & 62.6 \\ 
    & RIG-1                        & 65.20 & 63.00 & 63.08 & 65.20 & 64.06 & 59.00 & 60.21 & 64.06 \\ 
    & RIG-2                        & 67.19 & 63.19 & 62.57 & 67.19 & 65.79 & 61.18 & 59.79 & 65.79 \\ 
    & VDST-RL                       & 69.51 & 70.55 & 71.03 & 71.03 & 66.76 & 67.73 & 67.52 & 67.73 \\ 
    & \bf ADVSE-QGen                       &\bf 71.26 & \bf 72.73 & \bf 72.24 & \bf 72.73 & \bf 68.82 & \bf 69.88 & \bf 69.88 & \bf 69.88 \\ 
    \hline
     ADVSE-Guesser & \bf ADVSE-QGen                    &\bf 72.38 & \bf 73.59 & \bf 73.73 & \bf 73.73 & \bf 70.61 & \bf 71.10 & \bf 71.27 & \bf 71.27 \\ 
    \bottomrule
  \end{tabular}
\end{table*}

\subsubsection{Evaluation Metric and Comparison Models}

Following existing studies (such as \cite{de2017guesswhat}), we use the game success rate as the evaluation metric and evaluate in 3 generating way (i.e., sampling, greedy, and beam search (beam size=20)) by 2 test settings, i.e. New Object (games with seen images in train set but randomly sampled new target) and New Image (games with unseen images in test set).

We make comparisons in supervised training fashion and advanced training fashion (includes reinforcement learning and cooperative learning) respectively. The 3 supervised models are: the baseline SL \cite{de2017guesswhat}, the DM \cite{shekhar-etal-2018-ask} and the current state-of-the-art model VDST-SL \cite{Pang2019VisualDS}; 9 advanced training models are: baseline RL \cite{Strub2017EndtoendOO}, GDSE-C \cite{shekhar-etal-2019-beyond}, TPG \cite{zhao2018learning}, VQG \cite{Zhang_2018_ECCV}, ISM \cite{DBLP:journals/corr/abs-1812-06398}, Bayesian \cite{Abbasnejad2018WhatsTK}, RIG as rewards (RIG-1), RIG as a loss with 0-1 rewards (RIG-2) \cite{shukla-etal-2019-ask} and the current state-of-the-art model VDST-RL \cite{Pang2019VisualDS}.

\subsubsection{Quantitative Results}

Table 1 shows the comparisons among models in supervised learning and reinforcement learning, respectively. To be fair, all models in comparisons use the standard Oracle and Guesser model in this part.

{\bfseries Supervised learning.} As in the upper part of Table 1, our ADVSE-QGen achieves the best performance. With the standard Guesser \cite{Strub2017EndtoendOO}, the model achieves the success rate of 50.66\% on New object and 47.03\% on New game, exceeding the state-of-the-art model VDST-SL in all settings. 

{\bfseries Reinforcement learning.} As can be seen in Table 1 (lower part), the success rate of our ADVSE-QGen is significantly better than the previous methods in any case. Even though we use a simple 0-1 reward, compared with other models that use more finely designed rewards, we still achieve better performance. For example, our model achieves 9.08 points of improvement on New game compared to the VQG model, which designs three fine-grained rewards. Compared with the RIG-1 model that uses informative reward, our model achieves a higher success rate of 5.82 points on New game and 7.53 points on New object. Compared with the current state-of-the-art model VDST-RL, we have improved the success rate in all aspects, gaining an absolute advantage of 2.15 points on New game. In summary, using the benchmark Guesser \cite{Strub2017EndtoendOO} as the training environment, our model has achieved a maximum success rate of 72.73\% on New object and 69.88\% on New game, and achieves the new state of the art.

\begin{figure}[t]
  \centering
  \includegraphics[width=0.7\linewidth,height=9.3cm]{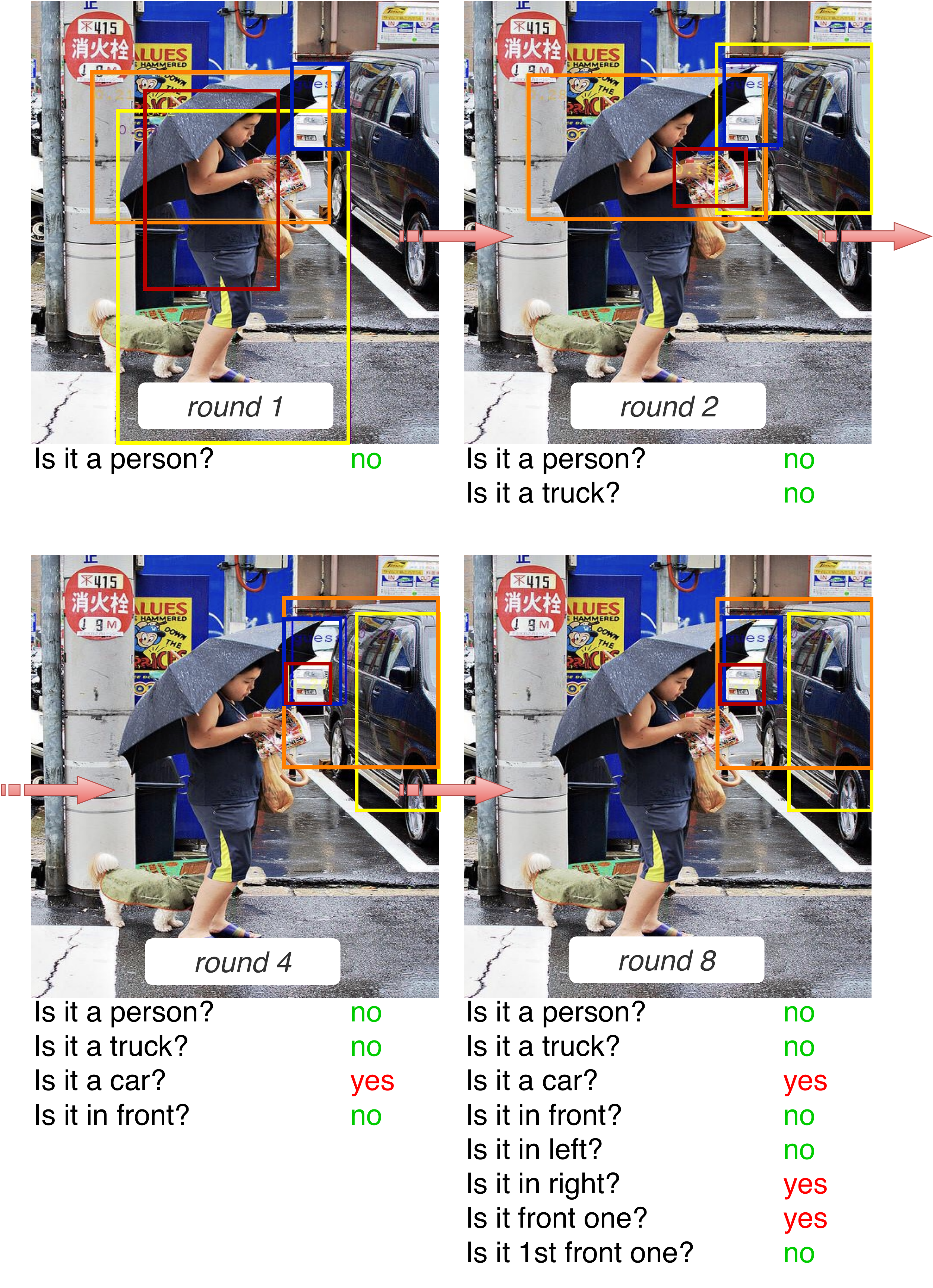}
  \caption{Illustration for the process of question generation.}
\end{figure}

\begin{figure*}[h]
\centering
\subfigure[Generated dialogue samples of the ADVSE-QGen trained by Supervised Learning.]{
\begin{minipage}[b]{0.9\textwidth}
\includegraphics[width=1\textwidth]{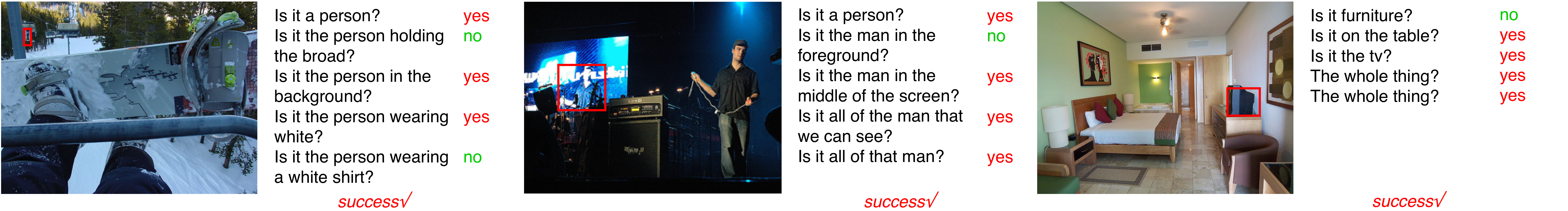}
\end{minipage}
}

\subfigure[Generated dialogue examples of ADVSE-QGen trained by Reinforcement Learning with the standard Guesser.]{
\begin{minipage}[b]{0.9\textwidth}
\includegraphics[width=1\textwidth]{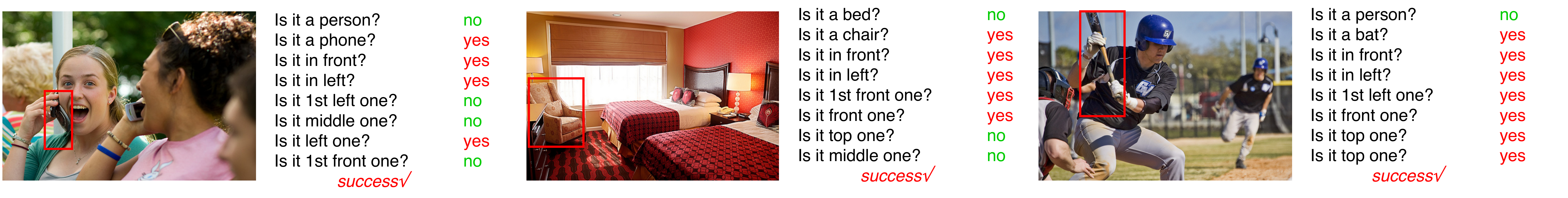}
\end{minipage}
}

\subfigure[Generated dialogue examples of the ADVSE-QGen trained by Reinforcement Learning with the ADVSE-Guesser.]{
\begin{minipage}[b]{0.9\textwidth}
\includegraphics[width=1\textwidth]{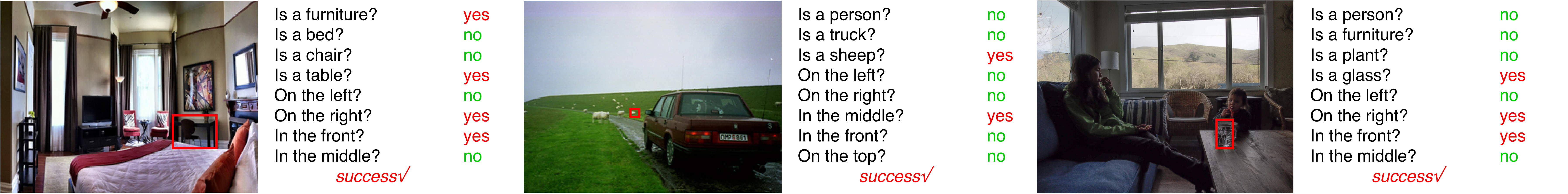}
\end{minipage}

}

\caption{Generated dialogue examples of the ADVSE-QGen under different training settings. The target is annotated in red.} \label{fig:1}

\end{figure*}

\subsubsection{Qualitative Results}

Figure 4 shows a visualized example of the question generation process and the changing of visual attention state ($att_t$) in our model. In each subgraph, the blue box annotates the target object; the red, orange and yellow boxes annotate the candidates with the top-3 largest attention weights at current round. As we can see, at the beginning of the conversation (round 1), the agent asks the \emph{"Is it a person?"}. After getting the answer of \emph{"no"}, the attention shifts to the non-persons and asks, \emph{"Is it a truck?"} (round 2). The Agent keeps on asking new objects until a positive answer to \emph{“Is it a car?”} is received, the attention is then focused on the differences among various cars, such as position, e.g. \emph{“Is it in front?”} is raised (round 4). Driven by following QA pairs, the attention state gradually focuses to the target object that is the more front-end car in the picture (round 8). It can be seen that the questions generated in each round are highly related to the current interested visual content, and the attention state changes according to the acquired answer. These phenomena fit well with the designed mechanisms.

Figure 5 gives additional dialogue examples generated by ADVSE-QGen under different training settings. As can be seen, the generated questions are highly related to the image. As the first example in Figure 5(a), the agent raises detailed questions, such as \emph{"Is it the person holding the board?"} and \emph{"Is it the person wearing white?"}, that describe the distinctive object feature comprehensively. Also, the ADVSE-based agents seem to follow some specific strategies. Notably, positive answers always bring about more detailed questions while negative answers lead to questions about the non-excluded objects. Moreover, the model is able to generate questions in a fine-grained differential style, such as \emph{"Is it the 1st front one?"}, which is very efficient for achieving goals.

\begin{table}[]
  \caption{Results of ablation study on QGen.}
  \begin{tabular}{lcccc}
    \toprule
    \multirow{3}{*}{ } & \multicolumn{2}{c}{\bf (\%)SL}      & \multicolumn{2}{c}{\bf (\%)RL}    \\
                          & \bf New     & \bf New     & \bf New    & \bf New \\ 
                          & \bf object & \bf game & \bf object & \bf game\\
    \midrule
    \bf ADVSE-QGen                & \bf 50.66 & \bf 47.03 & \bf 72.73 & \bf 69.88  \\ 
    w/o SO                        & 48.05 & 45.93  & 72.04 & 68.96  \\ 
    w/o ADFA                        & 47.69 & 45.21 & 70.48 & 68.20  \\ 
    w/o CVIF                        & 47.77 & 45.68  & 70.90 & 68.40  \\ 
    \bottomrule
  \end{tabular}
\end{table}

\subsubsection{Ablation Study}

We evaluate the individual contribution of the following components: 1) SO: we remove the Sharpening Operation (SO) in ADFA so that the question-guided attention is directly adjusted by the answer without polarizing afore; 2) ADFA: we remove the whole part of ADFA so that the attention is merely guided by history; 3) CVIF: we remove the whole part of CVIF so that only overall visual information can be used. We conduct the ablation study with the standard Oracle and Guesser.

As in Table 2, the result is showed in two training fashions, Supervised Learning (SL) and Reinforcement Learning (RL). It can be seen that without ADFA and CVIF, the performance of QGen model drops significantly, demonstrating their substantial contribution to goal-oriented visual question generation. Besides, the Sharpening Operation (SO) is validated to be an effective step in ADFA.

\subsection{Guesser}

\subsubsection{Training Details}

Guesser is trained in supervised way and is optimized by minimizing the negative likelihood loss. We use the Adam \cite{Kingma2014AdamAM} optimizer to train the Guesser model for 20 epochs with a learning rate of 1e-3, a batch size of 64. Learning rate is decayed by 0.9 per epoch. The hyperparameter $\gamma$ in Sharpening Operation in ADFA is set as 0.7.

\subsubsection{Evaluation Metric and Comparison Models}

Guesser model is evaluated by classification error rate. 
The 2 baseline models \cite{de2017guesswhat}: HRED, HRED-VGG, 3 attention-based models PLAN \cite{Zhuang2018ParallelAA}, A-ATT \cite{Deng2018VisualGV}, HACAN \cite{Yang2019MakingHM}, and 2 Feature-wise Linear Modulation (FiLM) models: single-hop FiLM \cite{perez2018film}, multi-hop FiLM \cite{strub2018visual}, are compared. 

\begin{table}[]
  \caption{Comparison Results of the Guesser.}
  \begin{tabular}{lc}
    \toprule
    \bf Model & \bf (\%)Test err       \\
    \midrule 
    HRED & 39.0     \\ 
    HRED+VGG & 39.6     \\ 
    PLAN & 36.6     \\ 
    A-ATT & 35.8     \\ 
    Single-hop FiLM & 35.7     \\ 
    Multi-hop FiLM & 35.0     \\ 
    HACAN & 34.1     \\ 
    \bf ADVSE-Guesser & \bf 33.15     \\
    \hline
    w/o SO &   33.45  \\ 
    w/o ADFA & 33.50    \\ 
    w/o CVIF & 33.65     \\ 
    \bottomrule
  \end{tabular}
\end{table}

\subsubsection{Quantitative Results}

Table 3 compares the test error of Guess models. Except for HRED (the first row in the table), all models utilize image feature, dialogue history, object spatial and category feature as input. As HRED+VGG compared to HRED, simply adding image feature will decrease the performance. However, applying appropriate attention mechanism to image helps the model to achieve higher performance, according to the PLAN, A-ATT and HACAN models. FiLM layers take effects either. Overall, it can be seen from the table that the Guesser model with our ADVSE structure achieves the lowest test error of 33.15\%, exceeds all the previous models and achieves the new state of the art.

\subsubsection{Qualitative Results}

\begin{figure}[t]
  \centering
  \includegraphics[width=0.7\linewidth]{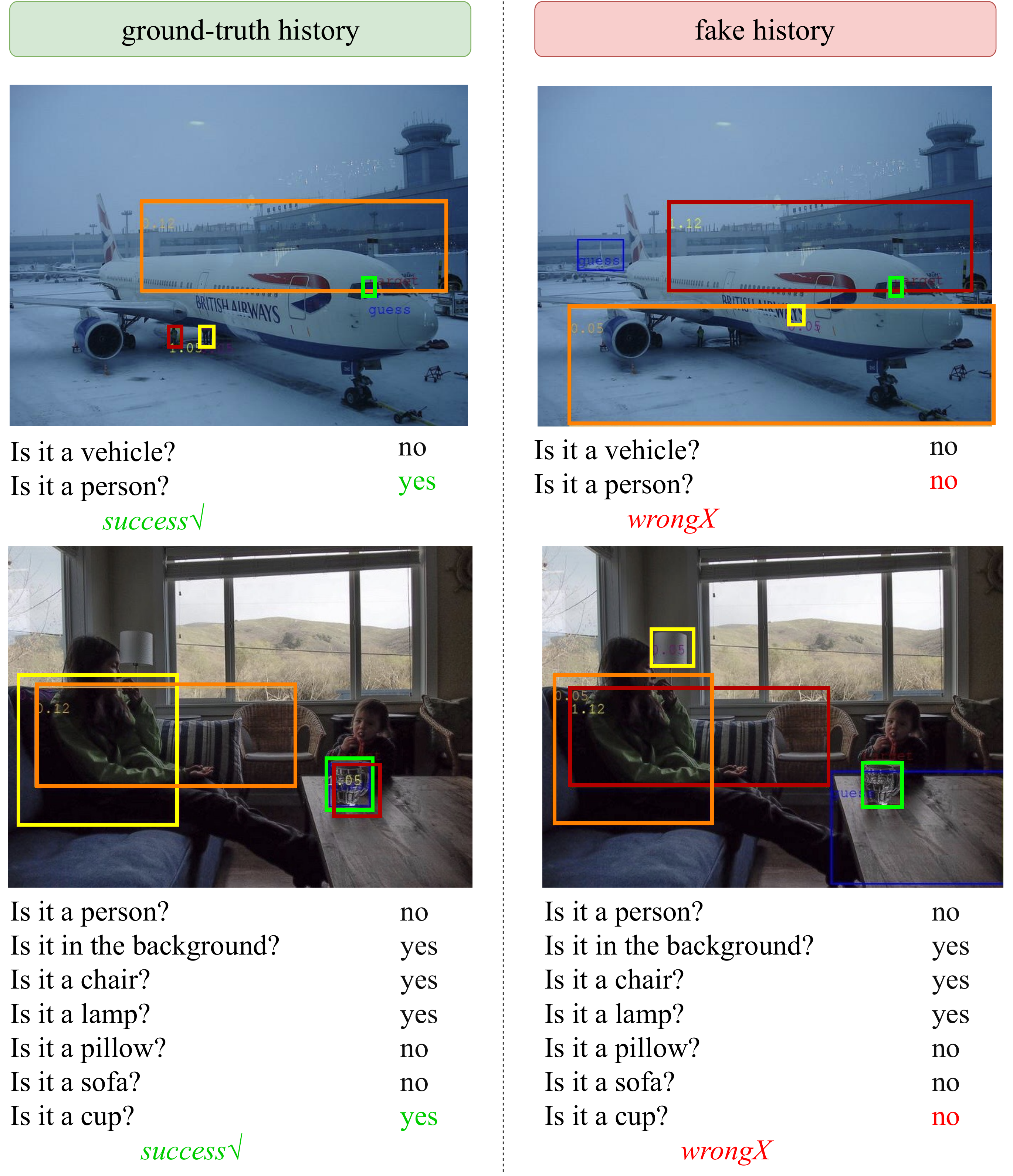}
  \caption{Visualization of the visual attention state in guess process. The left column is provided with the ground-truth history while the right column is provided with the fake history. The green box annotates the target object.
}
\end{figure}

Figure 6 illustrates the qualitative examples on Guess model. To illustrate the proposed ADFA mechanism, we visualize the visual attention state in the guess process. The red, orange and yellow boxes annotate the candidates with the top-3 largest attention weight. Further, we substitute the ground-truth history by fake history to make comparisons. It is clear that when current question is answered \emph{"yes"}, our guess model focuses on the question-relevant objects. On the contrary, as in fake examples, when current question is answered with \emph{“no”}, the model immediately transfers the attention to question-irrelevant objects. Moreover, as the right answers are taken place in the fake history, the guess results go wrong. The distinct results reflect the effectiveness of the proposed ADFA.

\subsubsection{Ablation Study}

We evaluate the individual contribution following the same setting as in section 5.1.5, i.e. SO, ADFA and CVIF. As in Table 3, without ADFA and CVIF, the Guesser results in comparatively worse performances. Besides, SO is still of significance in Guesser. To further illustrate the effect of each part, we conduct Significance Test on the four models. In concrete, we train each model for 10 times with random initialization and then conduct T-test on the collected data. Accordingly, ADFA, CVIF and SO are verified to be significant (with the p-value of 0.001, 0.001, 0.01).

\subsection{Joint QGen and Guesser}

Further, we combine the proposed QGen and Guess model. Both in the supervised learning and reinforcement learning processes for QGen, we replace the standard Guesser with our ADVSE-Guesser. We show the quantitative results Table 1. In SL, the model achieves the success rate of 54.06\% on New object and 50.94\% on New game, which are the best performances in supervised training to the best we know. In RL, the model achieves the success rate of 73.73\% on New object and 71.27\% on New game, which ulteriorly improves the performance. Overall, jointly using the ADVSE-QGen and ADVSE-Guesser, we achieve even better performance on GuessWhat?! task.

We give the generated dialogue examples in Figure 5(c). Jointly using ADVSE-QGen and ADVSE-Guesser generates dialogue in a more concise way. Still, the dialogue strategy is clear. Take the middle in Figure 5(c) as an instance. The agent firstly raises question to figure out the specific category of the target, like \emph{"Is a person?"}, \emph{"Is a truck?"}. Further, as obtained the positive answer \emph{"yes"} to \emph{"Is a sheep?"}, the agent then raises question in a detailed distinctive way to distinguish among many sheep. It successively asks \emph{"On the left?"}, \emph{"On the Right?"}, \emph{"In the middle?"}, \emph{"In the front?"} and finally reaches the target sheep, which is the middle but back one.

We combine the ADVSE-QGen and ADVSE-Guesser in a rather simple way in this section while further explorations for jointly using the two homologous models are expected in the future.

\section{Conclusions}
This paper proposes an Answer-Driven Visual State Estimator (ADVSE) to impose the significant effect of different answers on visual information in goal-oriented visual dialogue. First, we capture the answer-driven effect on visual attention by Answer-Driven Focusing Attention (ADFA), where whether to hold or shift the question-related visual attention is determined by different answer at each turn. Further, in Conditional Visual Information Fusion (CVIF), we provide two-types of visual information for different QA state and then conditionally fuse them as the estimation of visual state. Applying the proposed ADVSE to question generation task and guess task in Guesswhat?!, we achieve improved accuracy and qualitative results in comparison to existing state-of-the-art models on both tasks. Moving forward, we will further explore the potential improvements of jointly using the homologous ADVSE-QGen and ADVSE-Guesser.

\begin{acks}
We thank the reviewers for their comments and suggestions.  This paper is partially supported by NSFC (No. 61906018), MoE-CMCC “Artificial Intelligence” Project (No. MCM20190701), the Fundamental Research Funds for the Central Universities and Huawei Noah’s Ark Lab.
\end{acks}

\balance
\bibliographystyle{ACM-Reference-Format}
\bibliography{facif}

\end{document}